\documentclass[10pt,twocolumn,letterpaper]{article}

\usepackage[pagenumbers]{wacv} 

\definecolor{wacvblue}{rgb}{0.21,0.49,0.74}
\usepackage[pagebackref,breaklinks,colorlinks,allcolors=wacvblue]{hyperref}

\usepackage[table]{xcolor}

\def\wacvPaperID{1018} 
\def\confName{WACV}
\def\confYear{2026}

\title{Layout Anything: One Transformer for Universal Room Layout Estimation}

\author{Md Sohag Mia$^{1}$ \quad
        Muhammad Abdullah Adnan$^{2}$\\
$^{1}$Nanjing University of Information Science and Technology, Nanjing, China\\
$^{2}$Bangladesh University of Engineering and Technology, Dhaka, Bangladesh\\
{\tt\small shuvo2018@nuist.edu.cn}, {\tt\small adnan@cse.buet.ac.bd}
}

\begin{document}
\maketitle
\begin{abstract}

We present Layout Anything, a transformer-based framework for indoor layout estimation that adapts the OneFormer's universal segmentation architecture to geometric structure prediction. Our approach integrates OneFormer's task-conditioned queries and contrastive learning with two key modules: (1) a layout degeneration strategy that augments training data while preserving Manhattan-world constraints through topology-aware transformations, and (2) differentiable geometric losses that directly enforce planar consistency and sharp boundary predictions during training. By unifying these components in an end-to-end framework, the model eliminates complex post-processing pipelines while achieving high-speed inference at 114ms. Extensive experiments demonstrate state-of-the-art performance across standard benchmarks, with pixel error (PE) of 5.43\% and corner error (CE) of 4.02\% on the LSUN, PE of 7.04\% (CE 5.17\%) on the Hedau and PE of 4.03\% (CE 3.15\%) on the Matterport3D-Layout datasets. The framework's combination of geometric awareness and computational efficiency makes it particularly suitable for augmented reality applications and large-scale 3D scene reconstruction tasks.

\end{abstract}
    
\section{Introduction}

Indoor room layout estimation—the task of inferring the spatial structure of a room, including walls, floor, and ceiling, from a single RGB image—serves as a foundational pillar for scene understanding in computer vision. Its applications span augmented reality (AR), robotics navigation, virtual staging, and 3D scene reconstruction \cite{hedau1}. Early methods relied heavily on strong geometric priors like the Manhattan World assumption, leveraging vanishing points, edge features, and hypothesis ranking to recover cuboid layouts \cite{ramalingam2}. While effective in constrained settings, these approaches struggled with occlusions, clutter, and deviations from idealized geometries. The advent of deep learning catalyzed a paradigm shift, replacing handcrafted pipelines with data-driven models. Initial convolutional neural network (CNN)-based solutions focused on semantic segmentation of room planes or keypoint detection but often required complex, slow post-processing to refine layout hypotheses \cite{mallya3,dasgupta4}.

\begin{figure}[t] 
    \centering
    \includegraphics[width=1\linewidth]{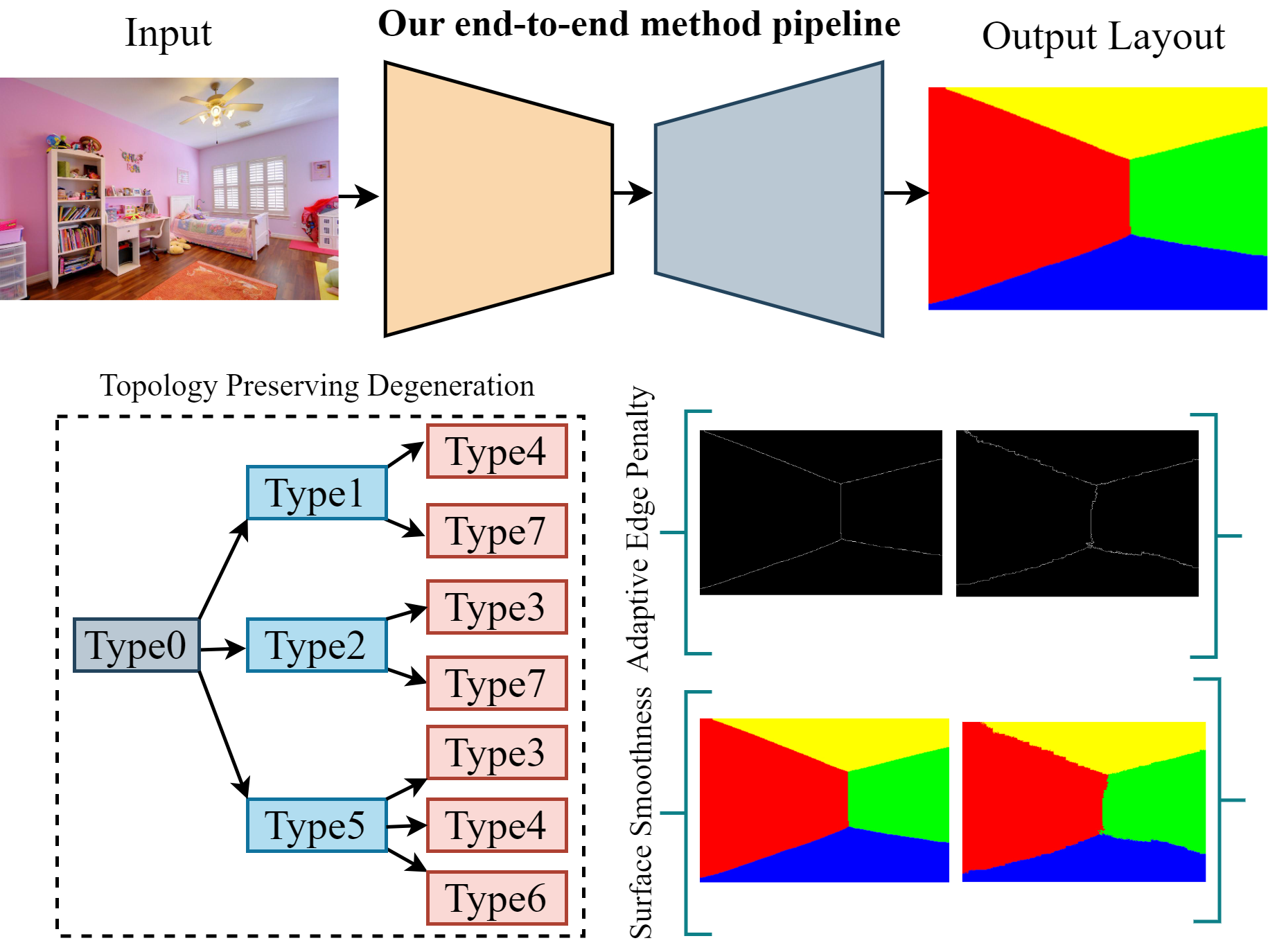} 
    \caption{Our end-to-end proposed method pipeline.}
    \label{fig:overview}
\end{figure}

Recent innovations have progressively relaxed restrictive assumptions and improved efficiency. End-to-end models like RoomNet demonstrated the feasibility of directly predicting ordered keypoints or segmentation masks using encoder-decoder architectures \cite{roomnet_lee16}. Subsequent works addressed limitations through diverse strategies: unsupervised learning with Spatial Transformer Networks (STNs) to warp canonical cuboids onto observed layouts (ST-RoomNet) \cite{st_roomnet5}; data augmentation via layout structure degeneration to handle class imbalance \cite{Lin6}; and the integration of 3D geometric reasoning through ordinal semantic segmentation and plane parameter regression \cite{IET7}. Concurrently, graph-based formulations emerged, parsing wireframes into junction-line graphs and employing heterogeneous graph transformers (HGTs) to infer polygonal plane projections, offering robustness to non-cuboid structures \cite{poly_hgt8}. Feature Pyramid Networks (FPNs) \cite{fpn23} were also adapted to preserve multi-scale features critical for detecting small keypoints efficiently, achieving real-time speeds while maintaining accuracy \cite{wang9}. Despite these advances, critical challenges persist. Methods balancing high accuracy (e.g., 3D-aware models \cite{IET7}) often incur significant computational costs from iterative optimization or clustering, hindering real-time deployment. Conversely, fast end-to-end models may sacrifice precision under severe occlusion or complex layouts \cite{wang9}. Furthermore, while graph-based approaches \cite{poly_hgt8} show promise in generalizing beyond cuboids, their performance remains sensitive to the quality of intermediate wireframe detections. This work bridges these gaps by introducing a novel, unified framework that synergizes the representational flexibility of graph neural networks with the efficiency of lightweight feature pyramids. Our approach directly infers geometrically consistent layouts without relying on cuboid priors or expensive post-processing, achieving high accuracy while operating at high speeds, as validated on benchmark datasets LSUN \cite{lsun10}, Matterport3D-Layout\cite{zgang_geolayout18} and Hedau \cite{hedau1}. The following sections detail our methodology, experimental validation, and contributions to advancing near real-time, generalizable room layout estimation.

Despite progress, key challenges remain: balancing accuracy and speed in cuboid or non-cuboid indoor scenes, handling severe occlusions where layout cues are hidden, and reducing the dependency on intermediate geometric representations such as vanishing lines or wireframe fitting. Our work (depicted in Figure~\ref{fig:overview}) addresses these challenges through a unified and efficient layout estimation pipeline tailored for practical applications in robotics, augmented reality, and indoor scene understanding. Specifically, we make the following key contributions:

\begin{itemize}
    \item We leverage a transformer-based segmentation framework, \textbf{OneFormer}, conditioned for layout-specific parsing using DiNAT-L, enabling precise surface-level predictions without relying on handcrafted cues or post-processing.
    
    \item We utilize \textbf{geometry-aware regularization losses} that explicitly encourage planar continuity, edge alignment, and boundary smoothness—essential for reliable structural reasoning in cluttered and occluded scenes.
    
    \item Finally, a \textbf{topology-preserving degeneration} strategy is utilized during training, which exposes the model to harder layout variations, improves generalization under occlusion, and enhances layout consistency in corner prediction.
\end{itemize}

Together, these components form an end-to-end trainable system that is both efficient and robust, making it well-suited for practical deployment in diverse application scenarios where accurate room layout is critical.

\section{Related Works}

\begin{figure*}[t] 
    \centering
    \includegraphics[width=1\linewidth]{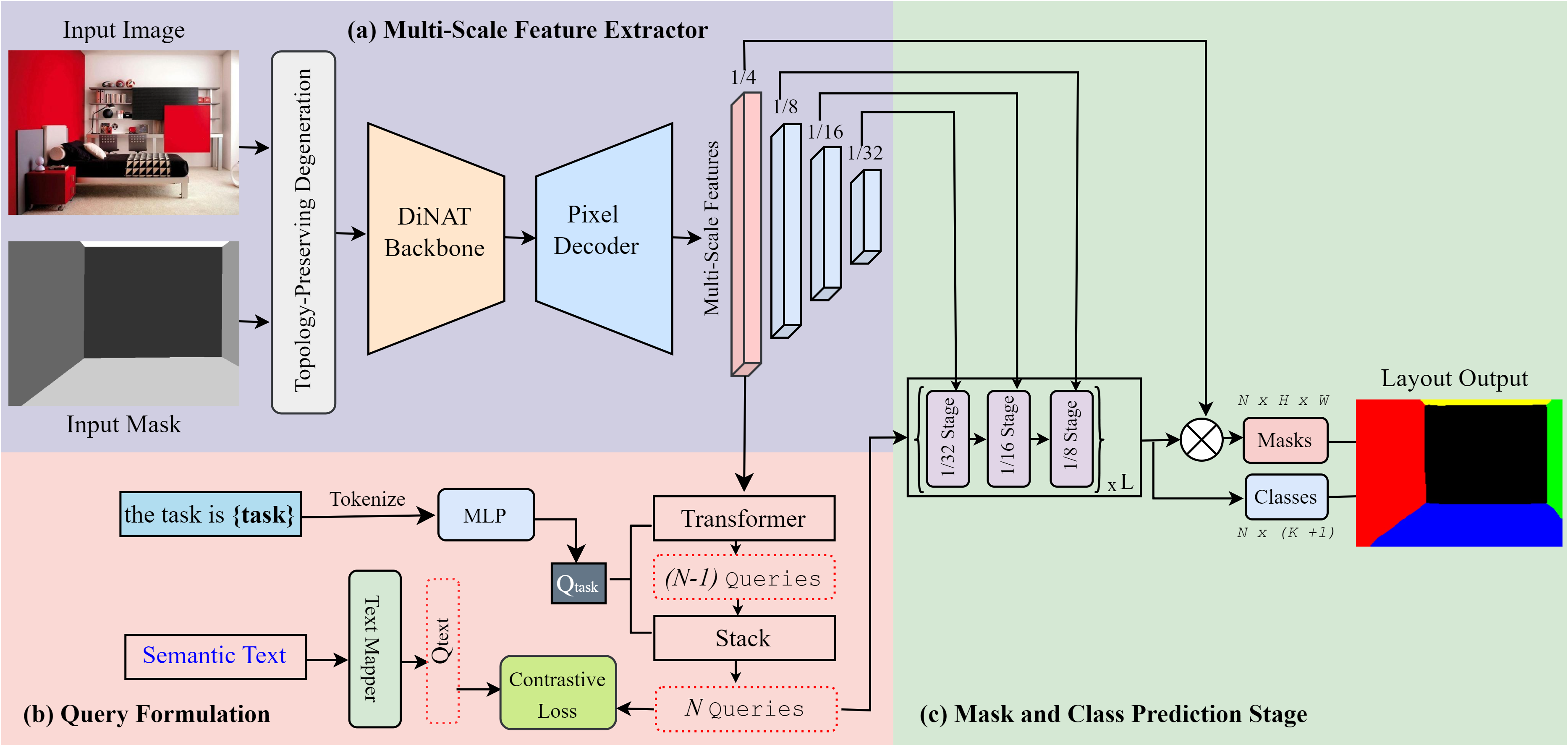} 
    \caption{Our Layout Anything Framework Architecture. (a) Multi-scale feature extraction using a DiNAT backbone and pixel decoder. (b) Task-conditioned query formulation with transformer-guided queries, task token ($\mathbf{Q}_{\text{task}}$), and contrastive learning. (c) Multi-stage transformer decoder producing task-dynamic class and mask predictions. Note: The input mask is only used during the training phase for applying the degeneration strategy. It is not required at test time.}
    \label{fig:model}
\end{figure*}

The evolution of room layout estimation reflects broader trends in computer vision, transitioning from geometric priors to data-driven learning. Early approaches leveraged strong assumptions about indoor scenes, particularly the Manhattan World hypothesis \cite{manhattan11}, which enforces orthogonal planes and cuboid structures. Hedau et al. \cite{hedau1} pioneered a three-stage pipeline: vanishing point detection, layout hypothesis generation, and ranking using handcrafted features. Subsequent refinements improved hypothesis sampling \cite{del_pero12} and optimization \cite{Schwing13}, but these methods remained brittle under occlusion and required significant computational overhead. The introduction of structured learning \cite{Nowozin14} marked an initial shift toward data-driven reasoning, though reliance on handcrafted features persisted.

Deep learning revolutionized the field through two dominant paradigms: \textbf{ semantic segmentation} and \textbf{Keypoint-based detection}. Fully Convolutional Networks (FCNs) enabled direct pixel-wise labeling of room planes (ceiling, walls, floor), as demonstrated by Dasgupta et al. \cite{dasgupta4}, who combined FCN predictions with vanishing point optimization. Mallya and Lazebnik \cite{mallya3} instead predicted "informative edges" to guide layout generation, while Ren et al. \cite{Ren15} adopted a coarse-to-fine FCN strategy with geometric constraints. These methods achieved accuracy gains but suffered from slow post-processing (e.g., 30s/frame \cite{dasgupta4}). Keypoint-based approaches like RoomNet \cite{roomnet_lee16} predicted corner coordinates and room type simultaneously, enabling end-to-end training but struggling with occluded corners. Extensions like Smart Hypothesis Generation (SHG) \cite{hirzer17} mitigated this by grouping room types by wall count, though multi-network inference increased complexity. Recent efforts focus on eliminating post-processing while enhancing generalization. Lin et al. \cite{Lin6} integrated edge-aware losses and layout-degeneration augmentation to handle class imbalance, achieving real-time inference. ST-RoomNet \cite{st_roomnet5} reimagined layout estimation as an unsupervised spatial transformation task, warping a canonical cuboid via predicted homography parameters. Feature Pyramid Networks (FPNs) were adapted by Wang et al. \cite{wang9} for keypoint detection, preserving point order during augmentation and refining corners to image boundaries—yielding state-of-the-art speed (31 FPS). Concurrently, 3D-aware methods emerged, transcending cuboid assumptions. Zhang et al. \cite{zgang_geolayout18} fused depth maps with plane parameters, while Yao et al. \cite{IET7} introduced ordinal segmentation to encode wall relationships and optimized layouts via depth-map intersection. Stekovic et al. \cite{stekovic19} leveraged render-and-compare for iterative 3D layout refinement, though at high computational cost.

Graph-based formulations represent the cutting edge. Gillisjo et al. \cite{poly_hgt8} parsed wireframes into junction-line graphs, using cycle sampling or Heterogeneous Graph Transformers (HGTs) to infer polygonal planes. This approach handled non-cuboid layouts but depended heavily on wireframe quality. Yang et al. \cite{yang20} fused plane and edge detections in a multi-branch network. Transformer architectures further pushed boundaries: PlaneTR \cite{tan21} used structure-guided attention for plane recovery, and ConvNext-based feature extractors \cite{liu_convnext22} enhanced spatial transformation robustness in ST-RoomNet.

Datasets and metrics have co-evolved with methodologies. Structured3D and InteriorNet-Layout \cite{IET7} introduced synthetic data with non-cuboid layouts and 3D annotations (depth, plane parameters), enabling training of complex models. Evaluation typically uses Pixel Error (PE), Corner Error (CE), and 3D corner distance \cite{IET7}, with recent additions like plane-based Average Precision.

\section{Methodology}
\label{sec:method}

This section presents our unified framework estimates of Manhattan-world room layouts by integrating task-conditioned segmentation with geometric priors, which synergistically combines the strengths of the OneFormer \cite{oneformer30} segmentation backbone with geometry-aware layout regularization inspired by classical cuboid layout models shown in Figure~\ref{fig:model}. We formulate layout estimation as a structured segmentation task, guided by planar surface semantics and reinforced by geometric and topological constraints. Below, we detail the core components of our architecture and training strategy.

\subsection{Task-Conditioned Joint Segmentation}

We adopt OneFormer as the backbone for semantic surface segmentation, leveraging its universal transformer-based architecture to model indoor environments with rich structural context. The segmentation task is conditioned using a natural language prompt such as ``the task is semantic,'' which is tokenized and embedded into a learnable task vector:
\begin{equation}
I_{\text{task}} = \text{``the task is } \{ \text{semantic} \} \text{''}, \quad Q_{\text{task}} = \text{Embed}(I_{\text{task}}).
\end{equation}

This task token is repeated $N-1$ times to initialize the object queries $Q' \in \mathbb{R}^{(N-1) \times d}$. The final set of object queries $Q \in \mathbb{R}^{N \times d}$ is:
\begin{equation}
Q = \text{Concat}(Q', Q_{\text{task}}),
\end{equation}
which are passed through a transformer decoder along with multi-scale image features $\{F_{1/4}, F_{1/8}, F_{1/16}, F_{1/32}\}$ to produce task-aware segmentation outputs.

To guide the learning of object-level semantics aligned with layout surfaces, we compute a bidirectional query-text contrastive loss:
\begin{align}
\mathcal{L}_{Q \rightarrow Q_{\text{text}}} &= -\frac{1}{B} \sum_{i=1}^B \log \frac{\exp(q^{\text{obj}}_i \cdot q^{\text{txt}}_i / \tau)}{\sum_{j=1}^B \exp(q^{\text{obj}}_i \cdot q^{\text{txt}}_j / \tau)}, \\
\mathcal{L}_{Q_{\text{text}} \rightarrow Q} &= -\frac{1}{B} \sum_{i=1}^B \log \frac{\exp(q^{\text{txt}}_i \cdot q^{\text{obj}}_i / \tau)}{\sum_{j=1}^B \exp(q^{\text{txt}}_i \cdot q^{\text{obj}}_j / \tau)}, \\
\mathcal{L}_{\text{contrastive}} &= \mathcal{L}_{Q \rightarrow Q_{\text{text}}} + \mathcal{L}_{Q_{\text{text}} \rightarrow Q},
\end{align}
where $\tau$ is a learnable temperature scaling parameter and $B$ is the batch size.

\subsection{Transformer-Based Surface Parsing}

In the context of room layout estimation, the accurate parsing of spatial surfaces is critical to reconstructing the topology of indoor environments. Our approach employs a transformer-based parsing mechanism, where OneFormer decodes image features and produces segmentation masks corresponding to the planar surfaces of a room—namely, the \textit{floor}, \textit{ceiling}, \textit{left wall}, \textit{right wall}, and \textit{front wall}. These are the five key semantic surfaces considered under the Manhattan World assumption.

Following multi-scale feature extraction and task-conditioned object query decoding, the transformer decoder produces $N$ object queries $Q \in \mathbb{R}^{N \times d}$, each enriched with both contextual visual cues and semantic intent. These queries are then used to generate per-class masks $\mathbf{M}^{(k)} \in \mathbb{R}^{H \times W}$ for each class $k \in \{1, \dots, K\}$, where $K=5$ represents the number of layout surfaces. Additionally, a $(K+1)^\text{th}$ class is reserved for the \textit{no-object} or background category.

Given the predicted set of masks, we assign each pixel $(i,j)$ the class label corresponding to the highest scoring class index across all predicted masks:
\begin{equation}
\hat{y}_{i,j} = \arg\max_k \mathbf{M}^{(k)}_{i,j}, \quad \text{where } \mathbf{M}^{(k)}_{i,j} \in \mathbb{R}^{H \times W}.
\end{equation}
Here, $\hat{y}_{i,j}$ is the predicted surface label at pixel $(i,j)$, selected from the maximum activation across the $K$ surface-specific binary masks.

We supervise segmentation using a weighted combination of cross-entropy, dice, and binary mask losses:
    \begin{equation}
    \mathcal{L}_{\text{seg}} = \frac{1}{HW} \sum_{i,j} \text{CE}(\hat{y}_{i,j}, y_{i,j}),
    \end{equation}

where $y_{i,j}$ is the ground truth class label at pixel $(i,j)$ and CE denotes the categorical cross-entropy. This loss ensures accurate per-pixel class assignments, which are essential for identifying distinct layout regions.

\begin{equation}
    \mathcal{L}_{\text{dice}} = 1 - \frac{2 \sum \hat{\mathbf{M}} \cdot \mathbf{M}_{\text{gt}}}{\sum \hat{\mathbf{M}}^2 + \sum \mathbf{M}_{\text{gt}}^2 + \epsilon}.
\end{equation}

Here, $\hat{\mathbf{M}}$ and $\mathbf{M}_{\text{gt}}$ denote the predicted and ground truth binary masks respectively, and $\epsilon$ is a small value to avoid division by zero. This term promotes spatial alignment and mask completeness, particularly useful for ensuring that walls and floors are captured as coherent planar regions.

\begin{equation}
    \mathcal{L}_{\text{bce}} = -\sum \left( \mathbf{M}_{\text{gt}} \log \hat{\mathbf{M}} + (1 - \mathbf{M}_{\text{gt}}) \log (1 - \hat{\mathbf{M}}) \right),
\end{equation}

which penalizes discrepancies in predicted pixel-wise probabilities. This loss encourages sharper foreground-background separation, which is especially beneficial near the surface boundaries.

The final supervision loss for surface parsing is a weighted combination of the three terms:
\begin{equation}
\mathcal{L}_{\text{surface}} = \lambda_1 \mathcal{L}_{\text{seg}} + \lambda_2 \mathcal{L}_{\text{dice}} + \lambda_3 \mathcal{L}_{\text{bce}},
\end{equation}

where $\lambda_1$, $\lambda_2$, and $\lambda_3$ are empirically chosen weights that balance the contribution of classification accuracy, spatial coherence, and pixel-wise confidence. In our experiments, we found $\lambda_1 = 2.0$, $\lambda_2 = 5.0$, and $\lambda_3 = 5.0$ to yield stable convergence and superior layout quality.

\begin{figure}[t] 
    \centering
    \includegraphics[width=1\linewidth]{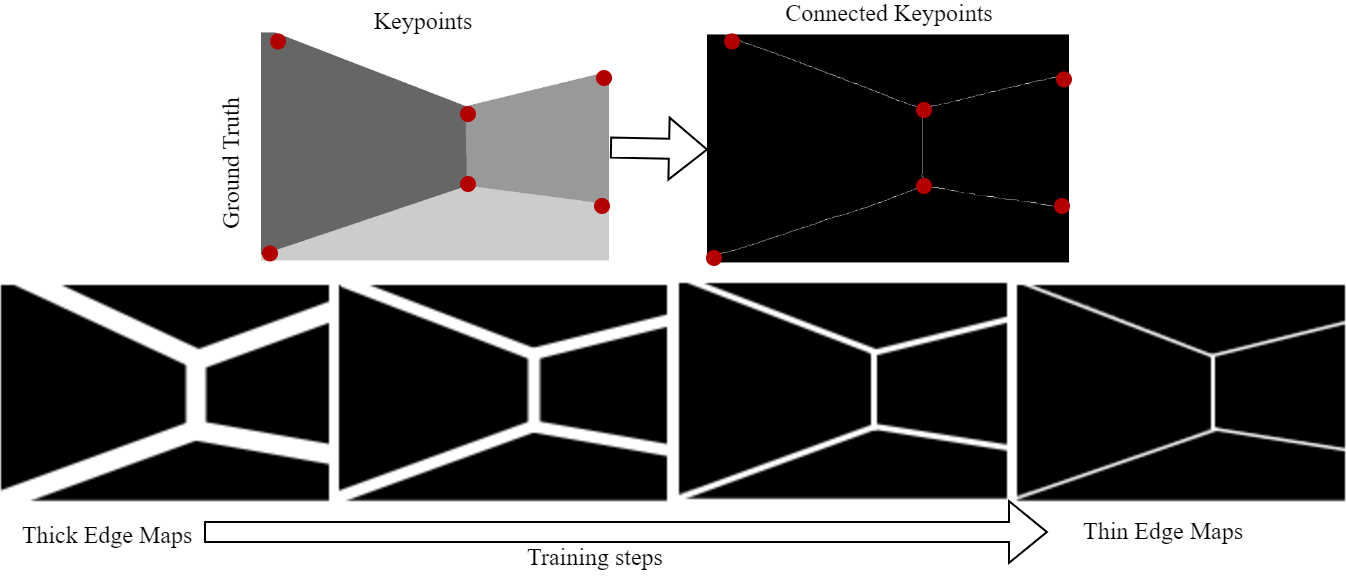} 
    \caption{Adaptive edge maps. (Top) Ground truth edge map generation. (Bottom) Edge maps with adaptive edge constraint applied.}
    \label{fig:edge}
\end{figure}

\begin{figure}[t] 
    \centering
    \includegraphics[width=1\linewidth]{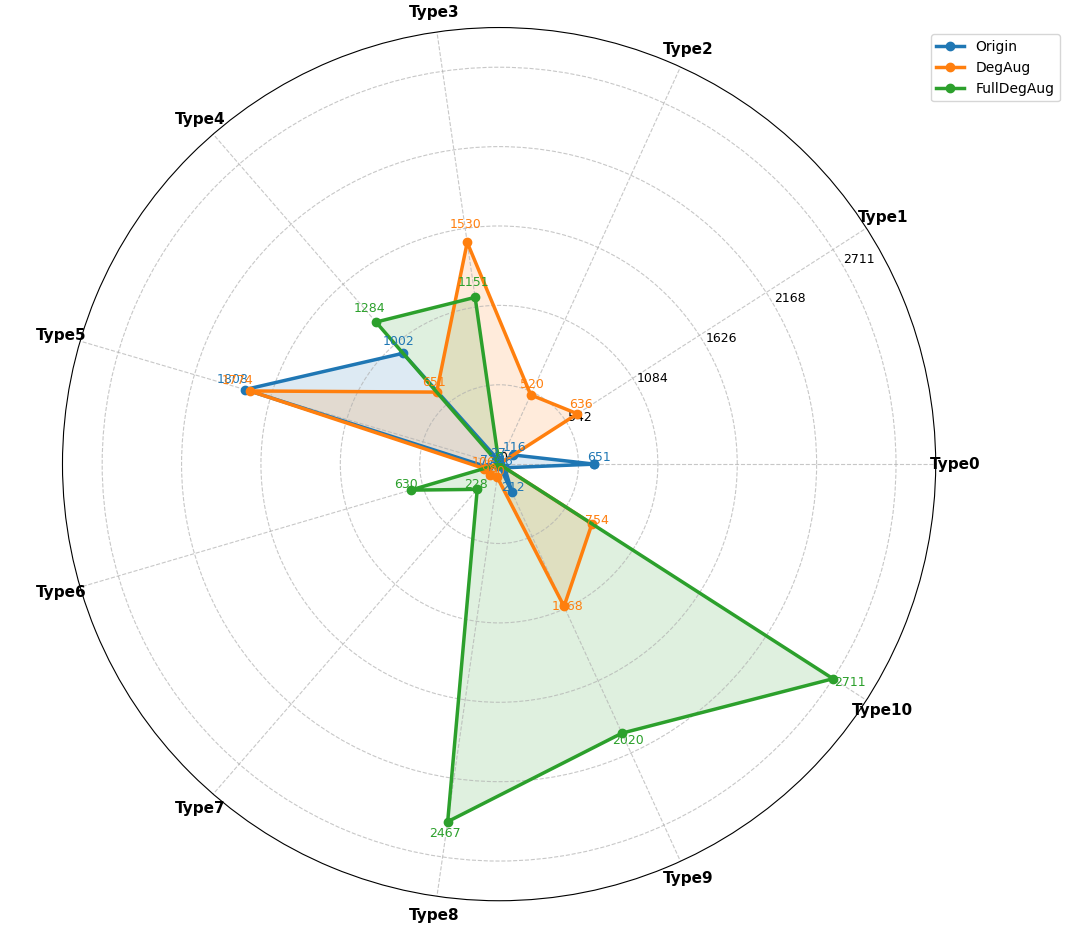} 
    \caption{The composition of room types within the LSUN Room Layout dataset.}
    \label{fig:lsun-deg}
\end{figure}


\subsection{Geometric Regularization Losses}
\label{ssec:geo_regularization}

To preserve structural integrity and reduce artifacts such as curved boundaries or noisy edges, we introduce geometry-aware loss functions tailored for planar layout estimation. The training strategy with these modules is shown in Figure~\ref{fig:edge}. Specifically, we employ two auxiliary constraints:

\paragraph{Adaptive Edge Penalty} promotes clean and linear surface transitions. Let $E_{\text{pred}}$ and $E_{\text{gt}}$ denote edge maps derived from gradient operations. The edge loss is computed as:
\begin{equation}
\mathcal{L}_{\text{edge}} = \text{BCE}(E_{\text{gt}}, 1 - \exp(-\|\nabla \mathbf{M}_{\text{pred}}\| / \sigma)),
\end{equation}
where $\sigma$ controls the edge tolerance and $\nabla$ denotes the spatial gradient of the predicted heatmap.

\paragraph{Surface Smoothness Loss} minimizes intra-plane variance, enforcing spatial coherence:
\begin{equation}
\mathcal{L}_{\text{smooth}} = \| \mathbf{M}_{\text{pred}} - \mathbf{M}_{\text{gt}} \|_2.
\end{equation}

These losses are combined into a geometric regularization term:
\begin{equation}
\mathcal{L}_{\text{geo}} = \lambda_4 \mathcal{L}_{\text{edge}} + \lambda_5 \mathcal{L}_{\text{smooth}}.
\end{equation}


\subsection{Topology-Preserving Degeneration}
\label{ssec:degenerative_aug}

Despite the high performance of the model on common room types, long-tail room configurations with fewer training samples can degrade layout generalization. To address this, we incorporate a novel \textbf{layout degeneration strategy} inspired by topological reasoning \cite{lsuntool31}, shown in Figure~\ref{fig:lsun-deg}. Given the hierarchical structure of room configurations under the Manhattan World, complex layouts can be systematically degenerated into simpler topologies by removing occluded or rare surfaces.

Let $\mathcal{T}$ be the set of all room types and $\mathcal{G}$ be a DAG (Directed Acyclic Graph) representing degeneration relations. For a node $t_i \in \mathcal{T}$ with children $\{t_j\}$ under degeneration, we generate synthetic samples by selectively masking surfaces:
\begin{equation}
\mathbf{M}_{\text{deg}} = \mathbf{M}_{\text{full}} \odot \mathbf{1}_{\text{retain}},
\end{equation}
where $\mathbf{1}_{\text{retain}}$ is a binary mask indicating preserved surfaces.

For example, a complex 'L-shaped living-room' node can degenerate into a simpler 'standard bedroom' node within the DAG by algorithmically removing the non-essential partition wall surface.

This augmentation increases diversity while preserving layout topology. The final objective integrates all modules:
\begin{equation}
\mathcal{L}_{\text{total}} = \mathcal{L}_{\text{surface}} + \mathcal{L}_{\text{contrastive}} + \mathcal{L}_{\text{geo}}.
\end{equation}

This design allows our model to robustly infer room layouts in both common and rare topologies, while maintaining semantic and geometric consistency.


\section{Experiment}

\subsection{Experimental Settings}

\noindent\textbf{Datasets:}  We train our method on two standard datasets for room layout estimation: the LSUN Room Layout dataset and the Hedau dataset. The LSUN dataset includes 4,000 training images, along with 394 validation and 1,000 test images, while the Hedau dataset consists of 209 training images, supplemented by 53 validation and 105 test images. Additionally, to evaluate performance on non-cuboid layouts and 3D-aware metrics, we use the Matterport3D-Layout\cite{zgang_geolayout18} dataset, which contains 4,939 training, 456 validation, and 1,965 test images with pixel-level depth and layout annotations. To enhance scene diversity during training, we apply our layout degeneration augmentation, random brightness adjustments, contrast variations, and horizontal flipping. Since flipping affects wall orientation, we appropriately modify the labels for left and right walls to maintain consistency.

\noindent\textbf{Evaluation Metrics:} We evaluate our method using standard metrics: pixel accuracy (PA\% $\uparrow$), pixel error (PE\% $\downarrow$), corner error ($e_{\text{cor}}$ $\downarrow$), and inference speed . Corner error measures the mean Euclidean distance between predicted and ground-truth corners, while pixel error quantifies segmentation consistency. Lower values indicate better performance for error metrics. All metrics are computed using the LSUN toolkit \cite{lsuntool31}.

\subsection{Implementation Details}

Our method was implemented in PyTorch 1.10 and evaluated on a single NVIDIA RTX 4090 GPU. Input images were resized to $256 \times 256$ resolution with random color jitter augmentation ($ \text{brightness} = 0.2$, $\text{contrast} = 0.1$). Training was conducted for 20 epochs using the AdamW optimizer ($\beta_1 = 0.9$, $\beta_2 = 0.999$) with an initial learning rate of $10^{-4}$, cosine decay, and a batch size of 4. At inference time, the model achieved a throughput of 114 ms per image at $256 \times 256$ resolution.

\begin{figure}[t] 
    \centering
    \includegraphics[width=1\linewidth]{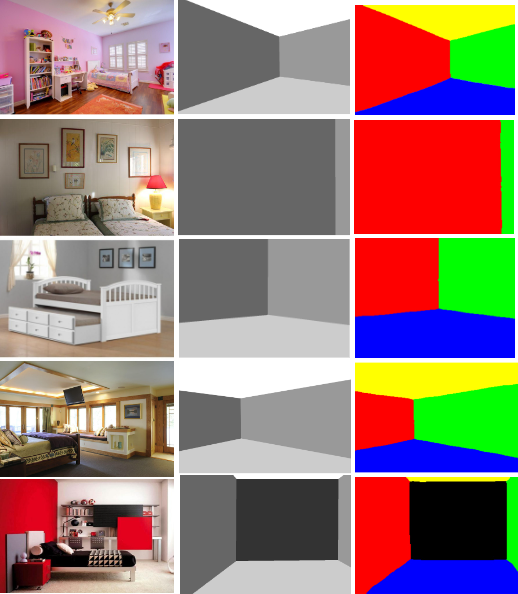} 
    \caption{Predicted results from our Layout Anything model.}
    \label{fig:predict}
\end{figure}

\subsection{Results on LSUN dataset}

To comprehensively evaluate the effectiveness of our proposed framework, we conduct experiments on the challenging LSUN Room Layout Estimation dataset and compare our results against a broad set of state-of-the-art methods spanning both traditional geometric approaches and recent deep learning-based models. We report four evaluation metrics: pixel accuracy (PA\%), pixel error (PE\%), corner error ($e_{\text{cor}}$), and inference speed measured in images per second, as summarized in Table~\ref{tab:lsun_results}.

Our model, Layout-Anything, achieves competitive, state-of-the-art performance across all core metrics, establishing a new benchmark on the LSUN dataset. Specifically, our method reaches a pixel accuracy (PA) of 94.57\%, surpassing strong baselines such as Lin et al.~\cite{Lin6} and Zheng et al.~\cite{zheng24}, which obtain 93.75\% and 93.05\%, respectively. In terms of pixel error (PE), our model maintains a low error of 5.43\%, on par with the best-performing models and significantly better than earlier CNN-based methods such as RoomNet~\cite{roomnet_lee16}, which reports 12.56\%. Most notably, our model achieves competitive corner error ($e_{\text{cor}}$) with a score of 4.02, outperforming all previous single-stage methods, including the prior best result by Zhao et al.~\cite{zhao25}, which had a corner error of 3.84 but at the cost of more complex multi-stage training and pretraining on a larger Sun-RGBD dataset. Accurate corner prediction is especially crucial for cuboid layout estimation, as it reflects the model’s capacity to preserve spatial geometry and infer accurate intersections of room surfaces. Our result in this metric highlights the strength of our simple and end-to-end architecture, which benefits from both transformer-based segmentation and geometric regularization. In addition to accuracy, our method offers practical advantages in terms of speed. With a throughput of 114ms image, Layout-Anything operates significantly faster than many previous models, including RoomNet (166ms), and Zheng et al. (170ms). This inference speed is achieved without sacrificing geometric fidelity or surface accuracy, making our model well-suited for efficient or interactive layout estimation tasks in downstream applications such as AR/VR, robot navigation, and indoor mapping. Compared to the original RoomNet~\cite{roomnet_lee16} and its re-implementation variant, which still suffer from higher PE and corner errors, our model demonstrates stronger generalization and robustness. Furthermore, earlier optimization-based methods such as those proposed by Hedau et al.~\cite{hedau1} and Dasgupta et al.~\cite{dasgupta4} yield substantially worse results in both PA and corner accuracy due to their reliance on hand-crafted geometric priors and vanishing point estimation, which are often brittle in cluttered or occluded indoor scenes. Predicted results from our proposed model are shown in Figure~\ref{fig:predict}. A quantitative comparison on the LSUN dataset with SOTA methods is shown in Table~\ref{tab:lsun_results}.

\begin{table*}[t]
\centering
\small
\begin{tabular}{l|cccccccccc}
\hline
Method & PA$\uparrow$  & PE$\downarrow$ & \textbf{$e_{cor}$}$\downarrow$ & $e_{3D\_cor}$$\downarrow$ & rms$\downarrow$ & rel$\downarrow$ & log10$\downarrow$ & ${\delta < 1.25}$$\uparrow$ & ${\delta < 1.25^2}$$\uparrow$ & ${\delta < 1.25^3}$$\uparrow$ \\
\hline
PlaneNet\cite{planenet} & 93.11 & 6.89 & 5.29 & 14.00 & 0.520 & 0.134 & 0.057 & 0.846 & 0.954 & 0.984 \\

GeoLayout-Plane\cite{zgang_geolayout18} & 94.16 & 5.84 & 4.71 & 12.05 & 0.448 & 0.109 & 0.046 & 0.891 & 0.973 & 0.993 \\

GeoLayout\cite{zgang_geolayout18} & 94.76 &  5.24 & 4.36 & 12.82 & 0.456 & 0.111 & 0.047 & 0.892 & 0.975 & 0.994 \\

\hline
\textbf{Layout-Anything} & \textbf{95.97} & \textbf{4.03} & \textbf{3.15} & \textbf{9.87} & \textbf{0.319 }& \textbf{0.071} & \textbf{0.031} & \textbf{0.967} & \textbf{0.991}  & \textbf{0.995} \\

\hline
\end{tabular}
\caption{Layout estimation results on the Matterport3D-Layout dataset.}
\label{matterport3dtable}
\end{table*}

\begin{table}[t]
  \setlength{\tabcolsep}{4pt}
  \centering
  \begin{tabular}{@{}lccrr@{}}
    \toprule
    Method &                  PA$\uparrow$  & PE$\downarrow$ &   $e_{cor.}$$\downarrow$ &  time$\downarrow$ \\
    \midrule
    Hedau \textit{et al.} \cite{hedau1}     & 75.77 &      24.23 &     15.48 &        \(-\) \\
    Mallya \textit{et al.} \cite{mallya3}    & 83.29 &      16.71 &     11.02 &         \(-\) \\
    Dasgupta \textit{et al.} \cite{dasgupta4} & 89.37 &      10.63 &     8.20 &          \(30 s\) \\
    Ren \textit{et al.} \cite{Ren15}       & 90.69 &      9.31 &      7.95 &          \(-\) \\
    Zhang \textit{et al.} \cite{zgang_geolayout18} &    87.51 &      6.58 &      \(-\) &          \( 150 s\) \\
    Zhao \textit{et al.} \cite{zhao25} &     94.71 &      5.29 &      3.84 &           \(-\) \\
    RoomNet \textit{et al.} \cite{roomnet_lee16} &   90.14 &      9.86 &      6.30 &           \(166 ms\)\\
    RoomNet (imp.)  \cite{roomnet_lee16}      &   87.44 &      12.56 &     \(-\) &           \(-\) \\
    Wange \textit{et al.} \cite{wang9} &    89.37 &      7.99 &      \(-\) &           \(32 ms\) \\
    Lin \textit{et al.} \cite{Lin6} &      93.75 &      6.25 &       \(-\) &           \(67 ms\) \\
    ST-RoomNet  \cite{st_roomnet5}           &    94.76 &      5.24 &      \(-\) &           \(102 ms\) \\
    Zheng \textit{et al.} \cite{zheng24}  &    93.05 &      6.95 &      \(-\) &           \(170 ms\) \\
    \midrule
     \rowcolor{gray!40}
    \textbf{Layout-Anything} &      94.57 &      5.43 &      4.02 &    \(114 ms\) \\
    \bottomrule
  \end{tabular}
  \caption{ Quantitative comparison on the LSUN dataset with SOTA methods.\protect\footnotemark[\value{footnote}]} 
  \label{tab:lsun_results}
\end{table}
\footnotetext{Speed comparisons are approximate and may vary by hardware and implementation. For reference: ST-RoomNet reported speeds on an RTX 3090 Ti at $400\times400$ resolution; our method runs at 114ms/image on an RTX 4090, Intel corei7@2.10GHz, and 32 RAM at $256\times256$ resolution.}


\begin{table}[t]
  \setlength{\tabcolsep}{4pt}
  \centering
  \begin{tabular}{@{}lccrr@{}}
    \toprule
     Method &                  PA$\uparrow$  & PE$\downarrow$ &   $e_{cor.}$$\downarrow$ &  time$\downarrow$ \\
    \midrule
    Hedau \textit{et al.} \cite{hedau1}     & 79.80 &      21.20 &     \(-\) &        \(-\) \\
    Mallya \textit{et al.} \cite{mallya3}    & 87.17 &      12.83 &     \(-\) &         \(-\) \\
    Dasgupta \textit{et al.} \cite{dasgupta4} & 90.28 &      9.72 &     \(-\) &          \(30 s\) \\
    Ren \textit{et al.} \cite{Ren15}       & 91.33 &      8.67 &      \(-\) &          \(-\) \\
    Zhang \textit{et al.} \cite{zgang_geolayout18} &    92.64 &      7.36 &      \(-\) &          \( 150 s\) \\
    Zhao \textit{et al.} \cite{zhao25} &     93.30 &      6.60 &      \(-\) &           \(-\) \\
    RoomNet \textit{et al.} \cite{roomnet_lee16} &   91.64 &      8.36 &      \(-\) &           \(166 ms\)\\
    RoomNet (imp.)  \cite{roomnet_lee16}      &   87.81 &      12.19&     \(-\) &           \(-\) \\
    Lin \textit{et al.} \cite{Lin6} &      92.59 &      7.41 &       \(-\) &           \(67 ms\) \\
    ST-RoomNet  \cite{st_roomnet5}           &    92.90 &      7.10 &      \(-\) &           \(102 ms\) \\
    Zheng \textit{et al.} \cite{zheng24}  &    92.79 &      7.21 &      \(-\) &           \(170 ms\) \\
    \midrule
    \rowcolor{gray!40}
    \textbf{Layout-Anything} &  92.96 &    7.04 & \textbf{5.17} &  \(114 ms\) \\
    \bottomrule
  \end{tabular}
  \caption{ Quantitative comparison on the Hedau dataset with SOTA methods.} 
  \label{tab:hedau_results}
\end{table}

\begin{figure}[t] 
    \centering
    \includegraphics[width=1\linewidth]{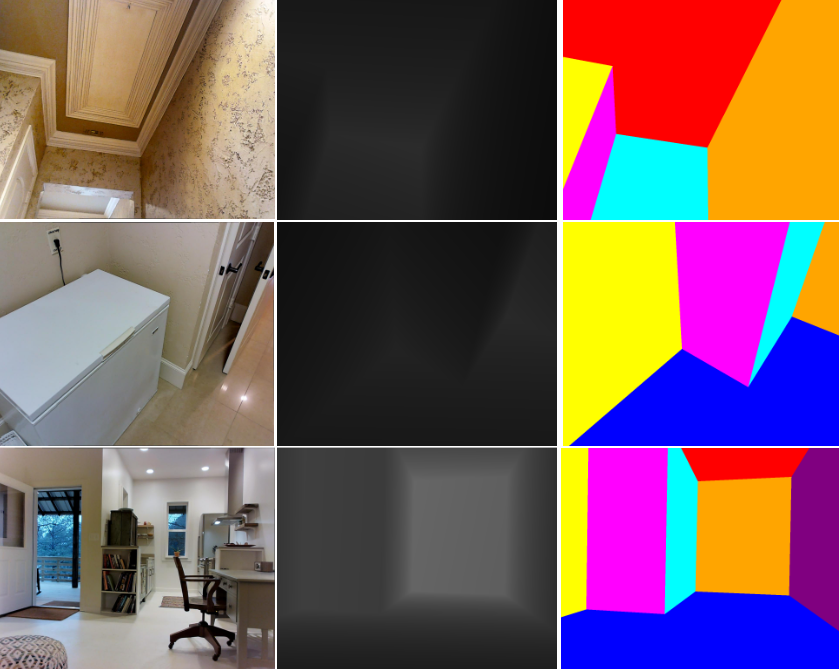} 
    \caption{Qualitative Results on Non-Cuboid Rooms from the Matterport3D Dataset.}
    \label{fig:matterport3d}
\end{figure}

\subsection{Results on Matterport3D-Layout dataset}

We further evaluate our model on the Matterport3D-Layout \cite{zgang_geolayout18} dataset, which includes both cuboid and non-cuboid room layouts with full 3D annotations. As shown in Table~\ref{matterport3dtable}, our Layout-Anything model achieves state-of-the-art performance across all metrics. It significantly outperforms previous methods in pixel accuracy (95.97\%), pixel error (4.03\%), and corner error (3.15\%). Moreover, our method excels in 3D geometric accuracy, achieving a 3D corner error of 9.87—a notable improvement over GeoLayout (12.82) and PlaneNet (14.00). Our model also demonstrates superior performance in depth prediction metrics, including RMS (0.319), relative error (0.071), and log10 error (0.031), and achieves the highest accuracy under all depth thresholds (\(\delta < 1.25\), \(1.25^2\), \(1.25^3\)). The qualitative results in Figure~\ref{fig:matterport3d} further confirm our model’s ability to accurately reconstruct complex room layouts, including non-cuboid structures, with sharp boundaries and consistent planar surfaces. These results highlight the effectiveness of our geometry-aware transformer architecture in generalizing to diverse and challenging indoor scenes.


\subsection{Results on Hedau dataset}

We also evaluate the generalization of Layout-Anything on the Hedau dataset. As shown in Table~\ref{tab:hedau_results}, our model achieves the second highest pixel accuracy (92.96\%) and the lowest pixel error (7.04\%), outperforming prior methods like ST-RoomNet~\cite{st_roomnet5} (92.90\%) and Lin et al.~\cite{Lin6} (92.59\%). We also report a corner error of 5.17\%, demonstrating accurate geometric inference. Our model runs at 114ms per image, faster than Zheng et al.~\cite{zheng24} (170ms) and ST-RoomNet (102ms), while heavy optimization methods like Zhang et al.~\cite{zgang_geolayout18} and Dasgupta et al.~\cite{dasgupta4} require seconds per image. These results confirm the robustness and efficiency of our approach for efficient or near real-time applications.

\subsection{Ablation Study}

In this subsection, we design a series of ablation studies to evaluate the effects of different modules of our Layout Anything model. 

\noindent\textbf{Ablation study on LSUN dataset:} We conduct an ablation study on the LSUN dataset to assess the individual contributions of our proposed components, namely the transformer backbone choice, geometric regularization, and topology-preserving degeneration. As shown in Table~\ref{tab:lsun_ablation}, we begin with a baseline using a vanilla FCN\cite{fcn26} model with a ResNet-101 \cite{resnet28} backbone and post-processing, which achieves a pixel error (PE) of 12.83\% and corner error ($e_{\text{corr}}$) of 8.36\%. Replacing this with the OneFormer segmentation model—without any post-processing—substantially improves performance. With ResNet-101 as the backbone, OneFormer \cite{oneformer30} reduces the PE to 8.30\% and $e_{\text{corr}}$ to 6.73\%, validating the effectiveness of task-aware query-based segmentation. Further experimentation with stronger backbones inside OneFormer shows continued improvement: ConvNeXt and Swin-L \cite{swin27} backbones reduce the PE to 7.92\% and 7.61\%, respectively. The best results among backbones are achieved with DiNAT-L \cite{Dinat29}, which yields a PE of 7.28\% and $e_{\text{corr}}$ of 5.98\%, while maintaining efficient performance at 114ms per image. Building upon this, we introduce a geometric regularization module that enforces edge alignment and planar smoothness, resulting in a notable reduction in PE to 6.74\% and corner error to 4.59\%, confirming that geometry-aware losses help produce more structured and coherent layouts. Finally, integrating the topology-preserving degeneration strategy further enhances generalization to diverse room configurations, improving PE to 5.43\% and achieving the lowest $e_{\text{corr}}$ of 4.02\%. Ablation study results are depicted in Table~\ref{tab:lsun_ablation}. Notably, these improvements are achieved without any additional inference time or post-processing.


\begin{table}[t]
  \setlength{\tabcolsep}{4pt}
  \centering
    \begin{tabular}{lcccc}
    \toprule
    \textbf{Model}            & PE$\downarrow$ & $e_{cor.}$$\downarrow$ & time$\downarrow$ & \textbf{PP} \\
    \midrule
    Vanilla FCN w/ResNet   &     12.83 &  8.36 & 168  &    YES \\
    \hline
    OneFormer w/ ResNet    &     8.30  &  6.73 & 137  &    NO \\
    OneFormer w/ Swin-L       &     7.61  &  6.05 & 103  &    NO \\
    OneFormer w/ ConvNeXt     &     7.92  &  6.12 & 129  &    NO \\
    
    \rowcolor{gray!30}
    OneFormer w/ DiNAT-L      &     7.28  &  5.98 & 114 &     NO \\
    \rowcolor{gray!30}
    + Geometric Losses &     6.74  &  4.59 & 114 &    NO \\
    \rowcolor{gray!30}
    + Degeneration            &     5.43  &  4.02 & 114 &     NO \\
    \bottomrule
  \end{tabular}
  \caption{Ablation study of the different components of our proposed Layout-Anything model on the LSUN dataset. (PP = Post Processing).}
  \label{tab:lsun_ablation}
\end{table}


\begin{table}[t]
  \setlength{\tabcolsep}{4pt}
  \centering
    \begin{tabular}{lcccc}
    \toprule
    \textbf{Model}            & PE$\downarrow$ & $e_{cor.}$$\downarrow$ & time$\downarrow$ & \textbf{PP} \\
    \midrule
    Vanilla FCN w/ResNet   &     12.28 &  \(-\) & 168  &    YES \\
    \hline
    OneFormer w/ ResNet    &     8.71  &  6.47 & 137  &    NO \\
    OneFormer w/ Swin-L       &     8.19  &  6.17 & 103  &    NO \\
    OneFormer w/ ConvNeXt     &     8.26  &  6.28 & 129  &    NO \\
    
    \rowcolor{gray!30}
    OneFormer w/ DiNAT-L      &     8.12  &  6.15 & 114 &     NO \\
    \rowcolor{gray!30}
    + Geometric Losses &    7.68  &  5.43 & 114 &     NO \\
    \rowcolor{gray!30}
    + Degeneration            &     7.04  &  5.17 & 114 &     NO \\
    \bottomrule
  \end{tabular}
  \caption{Ablation study of the different components of our proposed Layout-Anything model on the Hedau dataset. (PP = Post Processing).}
  \label{tab:hedau_ablation}
\end{table}


\begin{table}[t]
  \centering
    \begin{tabular}{lccc}
    \hline
    \textbf{Model}     & PE\%$\downarrow$ & $e_{cor.}$$\downarrow$ (\%) & PA\%$\uparrow$  \\
    \hline
    RoomNet\cite{roomnet_lee16}   &  12.56 &  7.10 & 87.44      \\
    w/ Geometric Losses          &   11.95  & 7.00 & 88.05      \\
    w/ Degeneration              &   10.12  &  6.47 & 89.88      \\
    \hline
  \end{tabular}
  \caption{Ablation study of our geometric losses and degeneration methods on LSUN dataset.}
  \label{tab:our-model-ablation}
\end{table}

\noindent\textbf{Ablation study on Hedau dataset:} We further validate our model's generalization on the Hedau dataset. As shown in Table~\ref{tab:hedau_ablation}, replacing the post-processed FCN+ResNet101 baseline (12.28\% PE) with OneFormer using the same backbone significantly improves performance (8.71\% PE), even without post-processing. Adopting stronger backbones yields additional gains: Swin-L and ConvNeXt achieve 8.19\% and 8.26\% PE, respectively, while DiNAT-L performs best (8.12\% PE, 6.15\% $e_{\text{corr}}$). Incorporating geometric regularization brings a notable improvement (7.68\% PE, 5.43\% $e_{\text{corr}}$), demonstrating enhanced boundary precision. Our degeneration strategy further refines accuracy, with the final model achieving the best results (7.04\% PE, 5.17\% $e_{\text{corr}}$) while maintaining efficient inference at 114ms per image. These results, summarized in Table~\ref{tab:hedau_ablation}, confirm that our complete framework—integrating OneFormer, DiNAT-L, geometric regularization, and degeneration—delivers superior accuracy and robustness for layout estimation.

\noindent\textbf{Ablation study of geometric losses and degeneration augmentation:} Our ablation study on the LSUN dataset demonstrates the effectiveness of our proposed geometric losses and degeneration augmentation when applied to the RoomNet framework (shown in Table~\ref{tab:our-model-ablation}). Adding geometric losses reduces RoomNet's PE from 12.56\% to 11.95\%, while incorporating degeneration further improves performance, achieving a PE of 10.12\% and a corner error of 6.47\%. These results confirm the generalizability of our modules across different architectural baselines.


\section{Limitations and Future Work}


Despite strong performance, our model occasionally produces imprecise boundaries in cluttered or partially occluded scenes, particularly when large furniture obscures layout cues, leading to under-segmentation or geometric inconsistencies. To address these issues, future work will integrate depth estimation and low-level edge cues to provide multimodal structural guidance. We also plan to explore uncertainty-aware estimation to explicitly model ambiguous regions and improve robustness in complex environments. These enhancements aim to deliver sharper and more reliable layouts under challenging conditions.

\section{Conclusion}

We presented Layout-Anything, a transformer-based model for robust indoor layout estimation from a single image. Our approach synergistically integrates a task-conditioned segmentation architecture with a novel topology-preserving degeneration strategy and geometry-aware losses. Extensive experiments demonstrate state-of-the-art performance across diverse datasets, including standard cuboid benchmarks (LSUN, Hedau) and the challenging Matterport3D-Layout dataset with non-cuboid rooms. The model achieves this with an efficient, single-stage pipeline that eliminates complex post-processing, making it suitable for practical applications in AR, robotics, and 3D reconstruction. Our work highlights the potential of combining learned segmentation priors with geometric reasoning for advanced scene understanding.

\section{Acknowledgement}
The project is funded by Basic Research Grant from Bangladesh University of Engineering and Technology (BUET).
{
    \small
    \bibliographystyle{ref}
    \bibliography{main}
}

\end{document}